\documentclass[conference]{IEEEtran}
\IEEEoverridecommandlockouts
\usepackage{amsmath,amssymb,amsfonts}
\usepackage{algorithmic}
\usepackage{graphicx}
\usepackage{textcomp}
\usepackage{xcolor}
\usepackage{enumitem}
\DeclareUnicodeCharacter{FFFD}{ }
\def\BibTeX{{\rm B\kern-.05em{\sc i\kern-.025em b}\kern-.08em
    T\kern-.1667em\lower.7ex\hbox{E}\kern-.125emX}}

\usepackage{biblatex}

\begin{document}

\title{Towards Data Governance of Frontier AI Models%
\thanks{Identify applicable funding agency here. If none, delete this.}
}

\author{
    \IEEEauthorblockN{1\textsuperscript{st} Jason Hausenloy}
    \IEEEauthorblockA{\textit{University of California, Berkeley}\\
    Berkeley, USA \\
    hausenloy@berkeley.edu}
    \and
    \IEEEauthorblockN{2\textsuperscript{nd} Duncan McClements}
    \IEEEauthorblockA{\textit{University of Cambridge}\\
    Cambridge, UK \\
    dm2020@cam.ac.uk}
    \and
    \IEEEauthorblockN{3\textsuperscript{rd} Madhavendra Thakur}
    \IEEEauthorblockA{\textit{Independent}\\
    New York, USA \\
    mt3890@columbia.edu}
}


\maketitle



\begin{figure*}[!t]
\centering
\includegraphics[width=0.8\linewidth]{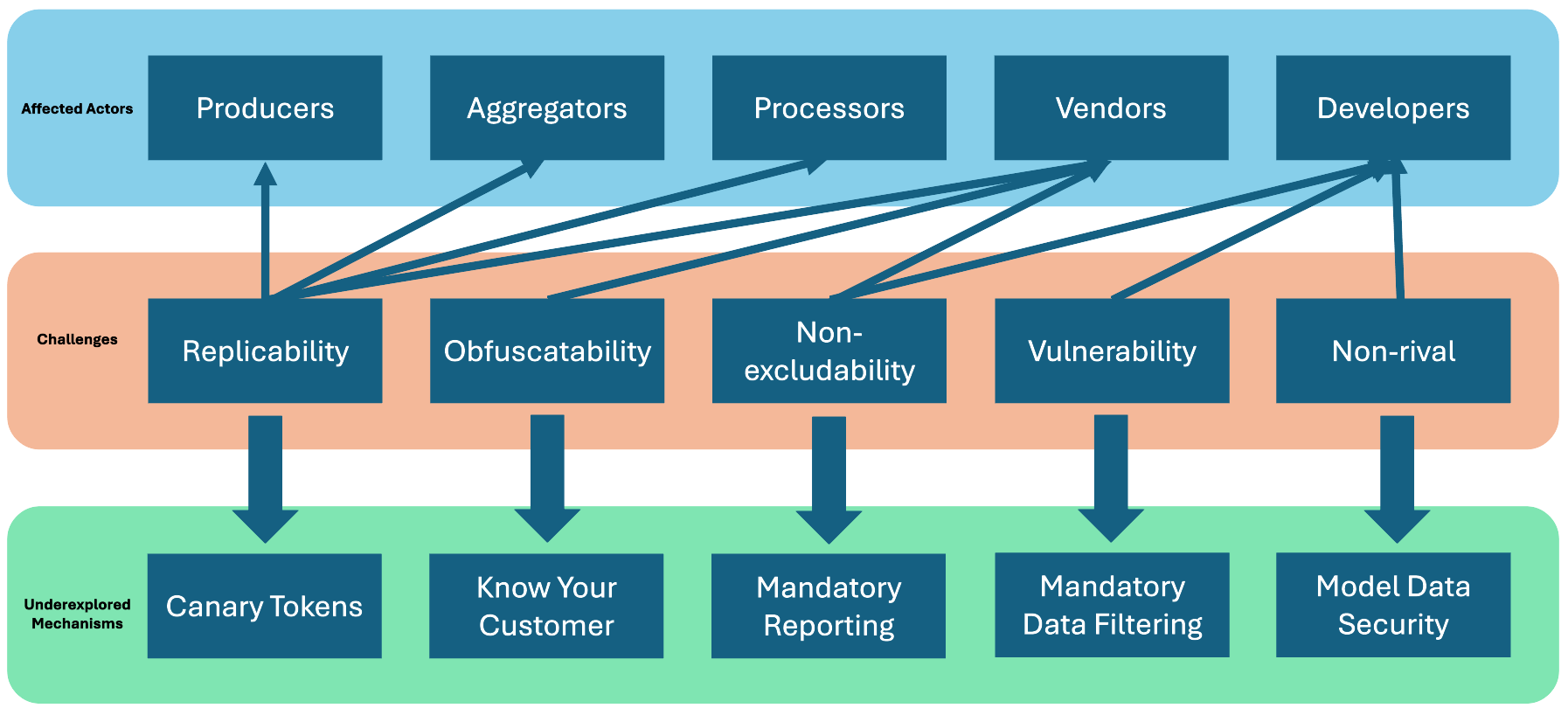} 
\caption{Proposed underexplored mechanisms in relation to challenges of data regulation and the AI Data Supply Chain}
\label{fig:combinedchart}
\end{figure*}

\begin{abstract}
Data is essential to train and fine-tune today’s frontier artificial intelligence (AI) models and to develop future ones. To date, academic, legal, and regulatory work has primarily addressed how data can directly harm consumers and creators, such as through privacy breaches, copyright infringements, and bias and discrimination. Our work, instead, focuses on the comparatively neglected question of how data can enable new governance capacities for frontier AI models. This approach for “frontier data governance” opens up new avenues for monitoring and mitigating risks from advanced AI models, particularly as they scale and acquire specific dangerous capabilities. Still, frontier data governance faces challenges that stem from the fundamental properties of data itself: data is non-rival, often non-excludable, easily replicable, and increasingly synthesizable. Despite these inherent difficulties, we propose a set of policy mechanisms targeting key actors along the data supply chain, including data producers, aggregators, model developers, and data vendors. We provide a brief overview of 15 governance mechanisms, of which we centrally introduce five, underexplored policy recommendations. These include developing canary tokens to detect unauthorized use for producers; (automated) data filtering to remove malicious content for pre-training and post-training datasets; mandatory dataset reporting requirements for developers and vendors; improved security for datasets and data generation algorithms; and “know-your-customer” requirements for vendors. By considering data not just as a source of potential harm, but as a critical governance lever, this work aims to equip policymakers with a new tool for the governance and regulation of frontier AI models.
\end{abstract}


\section{Introduction}


The development of today's frontier artificial intelligence (AI) models, highly capable foundation models, is inextricably linked to data, so much so that the systems are regularly defined by their training on ``broad data at scale'' \cite{anderljung2023frontierairegulationmanaging, bommasani2022opportunitiesrisksfoundationmodels}. There is a growing scientific consensus that, as well as tremendous benefit, such models may pose risks to public safety \cite{CAIS_AI_Risk_Statement, Bengio2024}. Yet, because of the rapid pace of AI development and the growing secrecy surrounding frontier model training, the production, aggregation, and processing of the datasets used by frontier models has thus far received little regulatory and public attention. As a key input to the pre-training and fine-tuning of models, we hope to demonstrate that governing data for frontier models, ``frontier data governance'', can be a promising approach to monitor and mitigate the risks as these models advance. 

While our primary focus is on catastrophic risks from frontier models, we acknowledge that data governance must address both current and future challenges. Present concerns including intellectual property violations \cite{scarcellaNvidiaMicrosoftHit2024}, algorithmic bias \cite{schwartzStandardIdentifyingManaging2022}, and model collapse \cite{shumailov2023curse} demonstrate immediate applications for data governance frameworks. These existing challenges provide valuable testing grounds for governance mechanisms that may later be critical for addressing catastrophic risks.

In particular, we focus on how policymakers can use the unique mechanisms within this approach to prevent the acquisition of specific dangerous capabilities, whether caused by malicious actors and potential misalignment. Existing ``data governance'' efforts emphasise detecting and preventing \textit{direct harms} arising from the abuse or misuse of data on consumers and creators. Examples of such efforts include regulation to protect individual privacy \cite{GeneralDataProtection, CaliforniaConsumerPrivacy}, data producer's copyright \cite{Directive-96-9-EC, Digital-Millennium-Copyright-Act}, and nondiscrimination in the workplace \cite{houseExecutiveOrderSafe2023}.
\footnote{Note: To limit the scope of our paper, we do not focus on the direct harms arising from data, though we recognise these cannot be completely disentangled. For example, if private companies can train on private data (like personal cloud drives or video call transcripts, as some sources have speculated), this could be prevented through enforcing existing privacy regulation – even if this would ultimately have ramifications for the size of datasets that model developers may need for further scaling \cite{Hale_2024, Zoom-data}. Likewise, some outlined mechanisms, such as canary tokens or data filtering, may be employed to prevent privacy or copyright-infringing datasets.} Instead, we focus on the comparatively neglected question of how data can enable new governance capacities for frontier AI models.

In this paper, we briefly overview previous attempts to govern data, and popular methods for regulating frontier AI models – namely ``compute governance'' and model evaluations. We then introduce ``frontier data governance'' as ``the policies, practices and mechanisms that monitor, regulate, and control data throughout the AI development pipeline, to mitigate risks and ensure responsible development of frontier AI systems.'' We then explain the importance of datasets to the acquisition of specific dangerous capabilities, and to the scaling towards larger, potentially more dangerous advanced AI systems. After, we show why the inherent properties that make  data so useful for training AI systems – it is non-rival, non-excludable and easily replicated – pose challenges for using it a governance lever, and that its current supply chain may leave it vulnerable to adversarial attacks, and obfuscatable from regulatory scrutiny. We then propose 16 mechanisms, of which we recommend five exploratory policies, and a single policy implementation:

\begin{enumerate}
    \item \textbf{Canary tokens:} data producers could embed unique identifiers in (particularly dangerous) data to detect and prevent unauthorized use in AI models
    \item \textbf{Mandatory data filtering:} model developers are required to implement a set of (automated) filtering processes to remove malicious or harmful content from training datasets
    \item \textbf{Mandatory reporting requirements:} after a certain threshold, model developers and data vendors must disclose their pre-training and fine-tuning datasets to a third-party evaluator.
    \item \textbf{Model data security:} model developers and data vendors should implement enhanced security measures to protect their datasets, and synthetic generation algorithms.
    \item \textbf{Know your customer regulations:} data vendors are required to collect, verify and disclose the identity of developers requesting certain datasets to the government.
\end{enumerate}

\section{Related work}

Governing data predates foundation models. In general, the term "data governance" has been used to refer to various mechanisms and techniques for the organizational handling of data \cite{benfeldtDataGovernanceCollective2020, plotkinDataStewardshipActionable2020}, and often in the context of AI \cite{janssenDataGovernanceOrganizing2020, vincent_data_leverage}. In particular, there is a rich literature on technical methods for privacy-preserving machine learning, including differential privacy, federated learning, and homomorphic encryption \cite{Xu_Baracaldo_Joshi_2021, Dwork_McSherry_Nissim_Smith_2006, McMahan_Moore_Ramage_Hampson_Arcas_2023, Homomorphic-encryption}.
Policymakers have translated some of these techniques into regulatory frameworks, with a wide focus on various \textit{direct harms} from data abuse and misuse, including privacy breaches, copyright infringement, algorithmic bias and discrimination \cite{GeneralDataProtection, CaliforniaConsumerPrivacy, Directive-2019/790}.

Recent work by \textcite{kunerMachineLearningPersonal2017} and \textcite{chandrasekaran_governance} provides a systematic framework for machine learning governance, highlighting data, model, and deployment governance cannot be performed independently. \textcite{janssenDataGovernanceOrganizing2020} propose a data leverage framework that demonstrates how data access can empower public oversight of technology companies. Our work builds on these foundations while specifically focusing on frontier model risks.

The relationship between training data and model behavior remains an active area of research. While initial work suggested direct correlations between training inputs and model outputs, recent studies suggest, beyond memorization and direct capability gain, emergent capabilities and behaviors may not clearly trace to specific training examples \cite{andonian2021emergent, wompLabs}. Our proposed governance mechanisms acknowledge this uncertainty while maintaining that data remains a crucial control point.

On the other end, the regulation of frontier AI systems has instead converged to two popular paradigms. First, the governance of computing power ("compute"), which has taken the form of export controls, compute thresholds, on-chip verification, among others\cite{sastryComputingPowerGovernance2024, CHIPS-act, heim2024trainingcomputethresholdsfeatures, brundage2020trustworthyaidevelopmentmechanisms}. Second, model evaluations, testing trained AI models before deployment, are a key component of the safety plans of leading model developers and proposed and implemented regulatory regimes such as licensing, third-party verification, auditing \cite{anthropic_responsible_scaling, phuong2024evaluatingfrontiermodelsdangerous}.

For governance, the early popularization of these two approaches is understandable. Compute is quantifiable and requires a highly-centralized supply chain, and evaluations generally provide a comprehensible indication of the lower bounds of a model's capability, to inform deployment decisions and development forecasts. \cite{sastryComputingPowerGovernance2024, phuong2024evaluatingfrontiermodelsdangerous}. Yet, there are still limitations. Once compute has been allocated (eg. when the data centers have been constructed), developers do not yet have the technology to reliably monitor their usage; algorithmic improvements mean that compute thresholds must be steadily lowered over time; evaluations can overlook emergent capabilities; correcting dangerous capabilities with existing techniques, such as reinforcement learning from human feedback, are still not perfect. \cite{casper2023openproblemsfundamentallimitations}

Recent work has demonstrated significant challenges in using evaluations alone to ensure model safety. For instance, \textcite{steinhardt2017certified} showed that even certified defenses against data poisoning can be circumvented. Similarly, work on model extraction \cite{carlini2021Extracting} and membership inference attacks \cite{duan2024membership} highlight the limitations of purely model-based approaches to governance. These findings further motivate frontier data governance as a complementary approach.

Our contribution lies at the intersection of data governance and frontier AI regulation, and best fits within the framework of the recent subfield of "technical AI governance." In particular, \textcite{reuel2024openproblemstechnicalai} first introduces the subfield, and provide a comprehensive taxonomy of open research questions in data, compute, algorithms, and deployment, sorted by governance capacity. On data, they primarily focuses on the data's ability to enable assessment, access, verification and security from a primarily technical lens, providing a comprehensive survey of progress in machine learning \cite{reuel2024openproblemstechnicalai}, while we seek to explicitly introduce the approach of "frontier data governance" and detail specific policy mechanisms to combat properties of data we have identified.


\section{Frontier data governance}

\subsection*{Motivation}

Data is a foundational input to AI models. Particularly for deep learning, models learn patterns, relationships and representations from the data they are trained on. The quality, quantity and nature of the training and fine-tuning data directly influence the model's capabilities, behaviors and potential risks. We define frontier data governance as "the policies, practices and mechanisms that monitor, regulate, and control data throughout the AI development pipeline, to mitigate risks and ensure responsible development of frontier AI systems." 

We briefly explore the risks from misuse and misalignment from frontier models, and data governance's role in mitigating these:

\begin{enumerate}
    \item \textbf{Misuse:} Frontier AI models have the potential to acquire capabilities that pose societal-scale risks, such as for bioweapon design, cyber attacks or autonomous weapons. Malicious actors could intentionally elicit or fine-tune these capabilities, training AIs on specialized datasets \cite{halawiCovertMaliciousFinetuning2024a, shu2023exploitability}. \textbf{While the relationship between training data and model behavior is complex \cite{grosse2023studying}, data governance can help prevent the most egregious misuse cases and create accountability throughout the development pipeline.} Compared to compute governance, which directly regulates the resources not the harmful content learned, or model evaluations, which occur after training, data governance can address the root cause by eliminating harmful content that provides these abilities from the training process, and can be enforced at multiple points throughout the supply chain.
    
    \item \textbf{Scaling:} AI models exhibit emergent capabilities as they scale in size and complexity, often in unpredictable ways \cite{andonian2021emergent}. However, scaling requires not just compute but also vast amounts of high-quality data; optimal performance and capability growth depend on both compute and data. Data requirements increase approximately linearly  with compute \cite{hoffmann2022trainingcomputeoptimallargelanguage}. The availability of high-quality, diverse public data is finite. Estimates suggest that publicly available data suitable for training could be exhausted by 2026–2032 under current growth trajectories \cite{villalobos2024rundatalimitsllm}. Without enough new data, models can overfit, and additional compute yields diminishing returns \cite{muennighoff2024scaling, xue2024repeat}. \textbf{This scarcity of high-quality data creates natural control points for governance, even as synthetic data capabilities advance \cite{meloNextgenerationDeepLearning2022}.} By controlling data access, policymakers can prevent rapid, uncontrolled scaling that may outpace society's ability to manage associated risks, thereby aligning AI development with societal readiness.
    
    \item \textbf{Misalignment:} \textbf{Recent work has shown that training data plays a crucial role in model alignment \cite{panEffectsRewardMisspecification2022, skalseDefiningCharacterizingReward2022}. While filtering harmful content alone cannot guarantee alignment, proper data curation can help prevent the most severe forms of misalignment.} Frontier data governance may be able to align frontier models that could otherwise be misaligned with human values or intentions, reducing the likelihood of scenarios where control over these systems is compromised. Filtering out harmful, biased, or malicious content from training datasets may reduce the likelihood of models learning undesirable behaviors. If the training distribution reflects the ethical standards and societal values may help align AI behaviors. 
\end{enumerate}

Beyond catastrophic risks, data governance can help address current challenges facing AI development. Recent lawsuits highlight issues with intellectual property rights \cite{scarcellaNvidiaMicrosoftHit2024}, while research demonstrates persistent problems with bias \cite{schwartzStandardIdentifyingManaging2022} and potential model collapse from unmonitored inputs \cite{bianchi_stereotypes, ghosh_gender_bias, shumailov_collapse}. These immediate concerns provide concrete test cases for developing and refining data governance mechanisms.
\textbf{
}Furthermore, even operationalizing concepts of 'malicious, harmful, or unsafe' content. Different stakeholders may have valid but conflicting perspectives on these definitions. For example, while developers might wish to filter harmful (eg. racist) content from training data, models may need exposure to such content to effectively detect and counter racism. Rather than prescribing universal definitions, our framework proposes mechanisms for making these decisions transparent and accountable through mandatory reporting requirements and public oversight.

Overall, frontier data governance may complement existing strategies, filling the gaps left by compute governance and model evaluations, and help proactively shape AI development, prevent dangerous capabilities, control scaling and even enhance alignment.

\subsection*{The AI data supply chain}

Frontier models rely on diverse datasets at various stages of their development and deployment. The AI data supply chain involves multiple stages: \textbf{production} (creation of raw data by users, creators, researchers, and organizations), \textbf{aggregation} (gathering data through web scraping and purchases by tech giants and aggregators), \textbf{processing} (cleaning and structuring data by AI company teams and academic institutions), \textbf{pre-training} (using these large datasets to optimize model parameters, done by AI companies and research labs), \textbf{fine-tuning} (adapting models for specific tasks with smaller datasets, involving specialized providers), \textbf{retrieval} (optional, incorporating external knowledge during inference, often with content partners), and \textbf{evaluation} (assessing model performance using curated datasets, conducted by internal teams, auditors, and researchers).

\begin{table*}[!t]
\caption{Actors in the AI Data Supply Chain \cite{teamNISTAIRCApp, yuHowDataShape2021}}
\centering
\resizebox{0.8\textwidth}{!}{
\begin{tabular}{|p{0.25\textwidth}|p{0.6\textwidth}|}
\hline
\textbf{Actor} & \textbf{Description and Examples} \\
\hline
Data Producers & \begin{itemize}
    \item Individual users (social media posters, video uploaders, bloggers)
    \item Content creators (YouTubers, podcast hosts, journalists)
    \item Researchers and academics (paper publishers, dataset sharers)
    \item Businesses and organizations (report producers, website maintainers)
    \end{itemize} \\
\hline
Data Aggregators & \begin{itemize}
    \item Tech giants (Google, Meta, Microsoft)
    \item Specialized data aggregators (CommonCrawl, WebCorpus)
    \end{itemize} \\
\hline
Data Processors & \begin{itemize}
    \item In-house teams at AI companies (OpenAI, Google DeepMind, Anthropic)
    \item Data science departments at universities and research institutions
    \item Cloud service providers
    \end{itemize} \\
\hline
Model Developers & \begin{itemize}
    \item Dedicated AI companies (OpenAI, Google DeepMind, Anthropic, Cohere)
    \item Research divisions of tech giants (Microsoft Research, Meta AI)
    \item Academic labs (Stanford AI Lab, MIT CSAIL)
    \item Open-source communities (Hugging Face, EleutherAI)
    \end{itemize} \\
\hline
Data Vendors & \begin{itemize}
    \item Content partners (Shutterstock, Getty Images, TIME)
    \item Specialized vendors (Scale AI, Surge AI, Appen)
    \item Data marketplaces (Kaggle Datasets, AWS Data Exchange)
    \end{itemize} \\
\hline
Evaluators & \begin{itemize}
    \item Internal teams within AI companies
    \item Third-party auditors (METR, Apollo Research)
    \item Academic researchers
    \item Government agencies (British and American AI Safety Institute)
    \end{itemize} \\
\hline
\end{tabular}
}
\label{aiactors}
\end{table*}

 For a more comprehensive breakdown of the AI data supply chain stages and the key actors involved at each step, refer to Table~\ref{aiactors}.

\begin{figure*}[!t]
\centering
\includegraphics[width=0.8\linewidth]{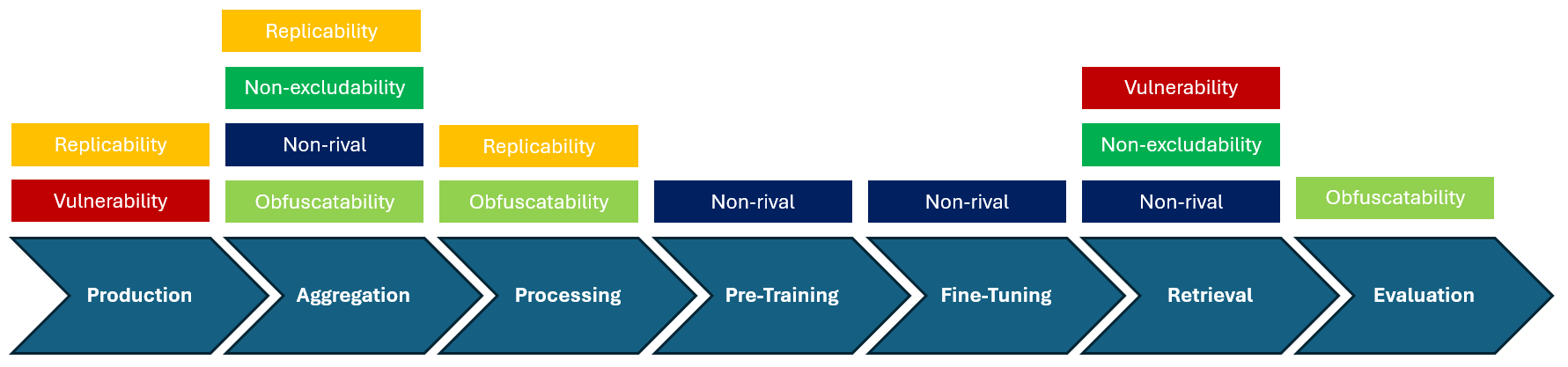} 
\caption{Challenges of regulating data across the data supply chain.}
\label{fig:challengeschart}
\end{figure*}

\section{Challenges}

While the inherent properties of data–non-rivalry, non-excludability, and replicability–make it incredibly useful for training AI systems, they also pose significant challenges for using data as a governance tool. Additionally, the current data supply chain may leave data vulnerable to adversarial attacks and obfuscation from regulatory scrutiny. Below, we explain why each of these properties causes a challenge.

\begin{enumerate}
    \item \textbf{Non-rivalry: }Data is a non-rivalrous good; one party's use does not diminish its availability or utility to others. While this has allowed for the widespread use and re-use of data in training AI models, this makes controlling or limits its use difficult \cite{duch-brownEconomicsOwnershipAccess2017, Jones_Tonetti_2019}.
    \item \textbf{Non-excludability:} Data is often non-excludable; it is difficult to prevent unauthorized access once available. Preventing unauthorized access to data, especially once it is available online, where malicious actors can obtain and use data without authorization, making it hard to control who uses data and for what purposes \cite{duch-brownEconomicsOwnershipAccess2017}. 
    \item \textbf{Replicability:} Data can be copied and replicated infinitely without degradation or significant cost. The ease of replicating data complicates efforts to control its distribution. Once data is shared or leaked, it can spread uncontrollably, making it nearly impossible to enforce restrictions or track all copies \cite{duch-brownEconomicsOwnershipAccess2017}.
\item \textbf{Vulnerability} (to adversarial attacks):  Data is susceptible to poisoning and extraction attacks, as adversaries can introduce malicious data into training datasets \cite{halawiCovertMaliciousFinetuning2024a, biggio2012poisoning, bagdasaryan2020backdoor}.\footnote{The barrier to entry for could be low. Public data sources can be manipulated with relatively few resources and technical expertise. Attackers can subtly alter data on websites or platforms that are commonly scraped for training data, embedding harmful content that could be ingested by model.} This can cause models to learn incorrect or harmful behaviors, leading to unpredictable or dangerous outcomes. Furthermore, using a variety of in-context data-based attacks, information can be inferred or extracted from trained models \cite{carlini2021Extracting}.  
\item \textbf{Obfuscatability:} Data can be acquired, transferred, and used in ways difficult to detect or monitor, often due to encryption, anonymization, or the use of covert channels. Unlike compute, which requires significant physical infrastructure (data centers, specialized hardware), data can be stored and transmitted using minimal hardware, such as on hard drives, and covert software-based approaches \cite{fu2016covert, nazari2020lightweightadaptablednschannel}, making it less visible to oversight mechanisms.
\end{enumerate}

These challenges are present across the AI Data Supply Chain, as seen in Figure 2. For each of these challenges, we highlight how our five proposed mechanisms can uniquely help combat them (in bold), and further classify the remaining 10 similarly.

\subsection*{Synthetic data}
The relationship between synthetic data and data governance presents several nuanced challenges. Even if a model's outputs are constrained through post-deployment safety mechanisms, the pre-deployment version of that model could still generate dangerous synthetic data during training. This risk exists because safety mechanisms like output filtering and RLHF are typically applied after pre-training \cite{casper2023openproblemsfundamentallimitations}, meaning pre-deployment versions could potentially generate harmful synthetic data that circumvents earlier data governance measures. While it follows logically that an AI system regulated to prevent certain capabilities would be unable to generate synthetic data yielding those same capabilities, this constraint only applies to systems operating within the governance framework. The primary risk thus emerges from unregulated systems that could generate and distribute synthetic training data enabling dangerous capabilities in other models, highlighting the importance of comprehensive regulation and international coordination. Recent work on model collapse \cite{shumailov2023curse} further suggests additional risks from synthetic data in training pipelines, which our proposed reporting requirements and security measures address through mandatory documentation and safeguards. Although more data-efficient architectures \cite{wilsonFutureAIWill2019} may reduce reliance on large datasets, this trend reinforces the importance of monitoring and governing the quality and provenance of all training data, whether synthetic or organic.

Furthermore, emergent technologies and industry trends cast doubt on the potential for data governance to act as a sustainable lever for regulation of the frontier. Particularly, as model training dataset sizes approach the limit of human-generated data, model developers are exploring opportunities for pushing the frontier. Although currently not a viable alternative for organic data \cite{longLLMsDrivenSyntheticData2024}, in the long term, synthetic data could supplant organic data as the primary source of training data in future generations of frontier models. This poses a threat to regulation of the data supply chain, as it relies on the progression of data from data producers to model developers; synthetic data, on the other hand, can be covertly generated and used by model developers, thus potentially allowing for unregulatable and adversarial data production and use. Likewise, the development of more data efficient model architectures, which many industry experts have suggested is the future, could threaten frontier data governance \cite{wilsonFutureAIWill2019}, as by reducing models' reliance on data, sufficiently large, and possibly dangerous, datasets could be created covertly, evading many of the detection and visibility mechanisms of frontier data governance.

\section{Existing mechanisms}

\textbf{While numerous approaches to data governance exist within AI development and adjacent fields, we first survey mechanisms that have seen some implementation or exploration in AI contexts. These approaches provide important context and foundations for the novel mechanisms we propose later.
}

\begin{table*}[!t]
\caption{Summary of Existing Mechanisms}
\centering
\renewcommand{\arraystretch}{1.5}  
\resizebox{1\textwidth}{!}{
\begin{tabular}{|p{0.25\textwidth}|p{0.6\textwidth}|}
\hline
\textbf{Mechanism} & \textbf{Description} \\
\hline
\textbf{Restricting and monitoring fine-tuning access} & Limiting fine-tuning capabilities or requiring identification can prevent unauthorized or malicious modifications of AI models. \\
\hline
\textbf{Restricting access to dangerous datasets} & Regulating access to unsafe datasets through licensing, background checks, and other restrictions helps mitigate risks in AI systems. \\
\hline
\textbf{Output watermarking} & Watermarking AI outputs facilitates traceability, enabling the identification of models responsible for generating unsafe content. \\
\hline
\textbf{Mandatory attribution for web scrapers} & Requiring web scrapers to attribute data sources enhances transparency and accountability in dataset creation, reducing unsafe content spread. \\
\hline
\textbf{Auditing retrieval data} & Monitoring and filtering retrieval data prevents adversarial manipulation of AI systems and ensures the safety of generated outputs. \\
\hline
\textbf{Decentralized volunteer classification for safe datasets} & Public volunteers can classify datasets for safety, aiding the development of AI models trained on vetted data. \\
\hline
\textbf{Implementing automated classifiers on input and output} & Classifiers trained to detect attacks on input and output data add a layer of protection against adversarial manipulation of AI models. \\
\hline
\textbf{Differential data spread} & Emphasizing beneficial data during training improves AI behavior, promoting safer outcomes in model predictions. \\
\hline
\textbf{Refining regulations to cover new data types} & Updating privacy laws to cover AI-generated and synthetic data strengthens oversight of AI systems. \\
\hline
\textbf{Proof-of-training-data verification} & Verifying datasets used in model training ensures proper attribution and prevents the use of unauthorized data. \\
\hline
\end{tabular}
}
\label{tab:safety_mechanisms}
\end{table*}

\textbf{Restricting and monitoring fine-tuning access} (model developers; vulnerability). For highly sensitive or capable models, directly limiting access to fine-tuning capabilities (eg. from the API) or requiring identification can prevent unauthorized or malicious modifications of AI models. Data poses risks at all stages of the pipeline, including fine-tuning, where technical frameworks exist for the cheap and covert exploitation of these post-training methods to instantiate unsafe agents \cite{shu2023exploitability, halawiCovertMaliciousFinetuning2024a}. Additionally, fine-tuning can be exploited through model poisoning attacks, where adversaries subtly manipulate the model's behavior during the fine-tuning process \cite{bagdasaryan2020backdoor}. The democratization of fine-tuning holds serious AI safety risks: ``defenders,'' or people interested in protecting AI, tend to have more resources, so as costs fall, more additional ``attackers'' gain fine-tuning access than ``defenders'' \cite{chanHazardsIncreasinglyAccessible2023a}. Thus, monitoring and restricting fine-tuning access, such as through license requirements or background checks, would mitigate these risks by blocking adversaries from accessing such resources. This approach already has some regulatory backing, covered under ``unsafe post-training modifications'' in California’s SB1047 \cite{BillTextSB1047}, and is also addressed in the proposed European Union's AI Act \cite{EUAIACT2021}.

\textbf{Restricting access to dangerous datasets} (data collectors, data processors, data vendors; non-excludability). Unsafe or malicious datasets can be used to create unsafe AI systems at various stages of the data pipeline, from pre-training to fine-tuning \cite{shu2023exploitability, halawiCovertMaliciousFinetuning2024a}. Regulating upstream data sources can significantly reduce these downstream risks. Measures such as mandating licenses, conducting background checks, implementing paywalls for accessing potentially unsafe datasets, or restricting the public dissemination of particularly dangerous datasets can prevent untraceable adversarial access to these resources. Existing regulatory initiatives, such as the European Union's General Data Protection Regulation (GDPR) \cite{GeneralDataProtection} and the California Consumer Privacy Act (CCPA) \cite{CaliforniaConsumerPrivacy}, aim to protect sensitive data from unauthorized access. These initiatives should be extended to cover data that could lead to the development of unsafe AI systems. 

\textbf{Output watermarking} (model developers; replicability). Implementing watermarking techniques to attribute data and identify the source model responsible for generating specific outputs allows for better filtering and tracking of AI-generated content. Model output watermarking involves embedding undetectable and unremovable tracing data into the outputs generated by AI models \cite{hartung1999multimedia}. These are different from canary tokens as they are built into the outputs, implemented by developers, rather than the inputs, implemented by data producers. While traditionally used by developers to protect intellectual property rights \cite{liSurveyDeepNeural2021, rouhani2018deepsigns}, output watermarking holds significant regulatory potential, across different modalities (including text, image, audio, video). Mandating model watermarking would allow regulators to trace unsafe or malicious content back to the source models, facilitating the identification and regulation of unsafe AI systems. Recent advancements have focused on developing robust watermarking methods for generative models to ensure accountability and mitigate misuse \cite{adi2018turning, zhang2018protecting}. 

\textbf{Mandatory attribution for web scrapers} (data aggregators; vulnerability). Requiring data aggregators to attribute the sources of their scraped data through standardized meta-reporting ensures transparency and accountability in data collection practices. The opaque processes of data aggregators often make it difficult to trace unsafe or harmful content back to its original source once it is included in a web-scraped dataset. This lack of traceability allows such content to proliferate unchecked into other datasets, potentially leading to unsafe AI models trained on this data \cite{bender2021dangers}. Mandatory and standardized attribution would not only facilitate traceability but also hold web scrapers accountable, enabling regulators to enforce safe scraping practices and prevent the spread of harmful content \cite{mitchell2019model}.

\textbf{Auditing retrieval data} (model developers; vulnerability). Using retrieval techniques can significantly enhance AI performance, especially with larger context windows \cite{dong2022survey}. However, these methods can also be exploited to circumvent restrictions on user prompts and elicit harmful information. By providing a sufficiently large number of examples of undesirable behavior, malicious actors can manipulate the model to produce unsafe outputs \cite{cheng2024trojanrag}. Since this vulnerability relies solely on input data, adversaries can create and distribute packages that are easily replicable to obtain harmful information. To mitigate these risks, it is crucial to audit and monitor retrieval data. User inputs can be analyzed to detect the presence of harmful examples; such inputs can then be entirely refused, portions that fail safety tests can be removed from model responses, or accounts that repeatedly submit such prompts can be tracked and suspended if unsafe behavior continues \cite{wallace2020imitation}. Implementing robust content filtering and input validation mechanisms can further enhance security \cite{xu2020recipes}. Additionally, employing techniques like differential privacy during training can help prevent the leakage of sensitive information \cite{dwork2014algorithmic}.

\textbf{Decentralized volunteer classification for safe datasets} (data aggregators, data vendors, data processors; obfuscatable). To break out of the chicken-and-egg paradox of relying on AI models to assess the safety of datasets used to train themselves, we propose leveraging volunteer classification efforts. Projects like Galaxy Zoo, launched in 2007, demonstrate the effectiveness of citizen science, where public volunteers classified galaxies and provided valuable training data that eventually enabled machine learning models to automate these tasks \cite{fortson2012galaxy}. Similarly, volunteers could be enlisted to review random pieces of data and classify their safety according to specified criteria. Once sufficient labeled data is gathered from volunteers, a classifier could be trained to automate verification in the future, with periodic updates from volunteers as the relative safety of data evolves. This approach harnesses the collective intelligence of the public to enhance dataset safety, ensuring that AI models are trained on vetted data \cite{howe2006rise, kittur2008crowdsourcing}. Moreover, involving volunteers in data classification promotes transparency and public trust in AI systems \cite{franzoni2014crowd}. However, challenges such as ensuring annotation quality and protecting volunteers from exposure to harmful content must be carefully managed \cite{schmidt2017survey}.

\textbf{Implementing automated classifiers on input and output} (model developers; vulnerability). Fine-tuning a small language model to detect signs of data extraction and other common attacks can provide an additional layer of security on top of existing LLMs. Data-based input attack vectors are attractive to adversaries due to their accessibility; for instance, attacks on web-scraped datasets can be executed simply by injecting large amounts of data expressing misaligned intent \cite{wallace2021concealed}. Similarly, output attacks, such as data extraction, involve malicious prompting that exploits vulnerabilities in the data pipeline, potentially leading to unauthorized access to sensitive information \cite{carlini2021Extracting}. Implementing input/output classifiers can detect many of these common attacks, blocking rudimentary threats with relatively simple techniques \cite{inan2023llamaguardllmbasedinputoutput}. These classifiers act as gatekeepers, analyzing inputs and outputs for signs of malicious activity and preventing the exploitation of LLMs.

\textbf{Differential data spread} (data producers, data aggregators, data vendors, model trainers; vulnerability).  Implementing techniques to assign greater weighting to positive or beneficial data—or increasing the raw quantity of such data in datasets—can promote more favorable outcomes in AI models. For example, incorporating ``good stories'' about AGI or emphasizing ethical and aligned content can steer models toward safer behaviors. Meta's Llama 2 model demonstrates the possibility of intentional data weighting by model developers; their pre-training data mix was curated to contain specific proportions of content types, showing that certain types of content can be deliberately emphasized in the training data \cite{touvron2023llama}. Techniques like curriculum learning, where training data is presented in a meaningful order to improve learning efficiency and performance, support the efficacy of intentional data weighting \cite{bengio2009curriculum}. Beyond this, differential data spread could be executed through the large-scale creation of beneficial data by data producers, an increased sampling of beneficial data by data aggregators, such as through duplication of beneficial data, and the curation of datasets with high amounts of beneficial data by data vendors. 

\textbf{Refine existing regulations to include mechanisms of data generation and collection} (regulators, evaluators; synthetic data). Existing regulatory frameworks for data privacy are invaluable but need to evolve as methods of data collection and generation advance. For example, the EU's GDPR defines personal data but does not explicitly address synthetic or AI-generated data that can be linked back to individuals \cite{GeneralDataProtection}. Expanding the GDPR's definition to explicitly include synthetic and AI-generated data would extend regulatory oversight to these novel forms. Similarly, consent mechanisms mandated by regulations like the CCPA need to evolve beyond explicit data collection to address AI systems capable of inferring sensitive information from aggregated non-sensitive data, or to regulate non-sensitive data itself \cite{CaliforniaConsumerPrivacy}. Updating these privacy regulations is essential for the effective governance of emergent technologies like synthetic data and would enhance protections against potential misuse of AI systems. Incorporating provisions from recent legislative proposals, such as the EU's proposed Artificial Intelligence Act, which addresses AI-specific risks and governance, could further strengthen the regulatory framework \cite{EUAIACT2021}. Additionally, research demonstrating how AI models can infer personal information from seemingly anonymized data underscores the need for refined regulations \cite{Shokri2017Membership, carlini2021Extracting}.

\textbf{Proof-of-training-data verification} (model developers, regulators; non-excludability). Proof-of-training-data is an emerging mechanism for verifying the datasets used to train machine learning models \cite{choi2023toolsverifyingneuralmodels}. This mechanism holds significant potential as an enforcement tool for regulators, enabling the auditing of models to ensure correct dataset attribution and confirming that models have not been trained on prohibited or unauthorized datasets. For instance, if a malicious model developer circumvented the standard data pipeline to obtain a sensitive or dangerous dataset through illicit means, resulting in a harmful AI agent, regulators could employ Proof-of-training-data verification to determine whether the inaccessible dataset was used during training. Techniques such as data provenance tracking \cite{herschel2017provenance}, model watermarking \cite{uchida2017embedding}, and cryptographic verification methods \cite{ben2019scalable} can enhance this verification process by ensuring transparency, enabling ownership verification, and preserving data privacy, respectively. This approach addresses a critical regulatory challenge posed by black-box models, where outputs are often untraceable to specific training data, thereby establishing accountability measures for model developers even after training is complete.

\section{Underexplored mechanisms}

We highlight 5 novel governance applications of technical mechanisms that each address a central challenge to governing data and together form a united AI governance policy framework.

\subsection*{Canary tokens}

\begin{figure*}[!t]
\centering
\includegraphics[width=0.8\linewidth]{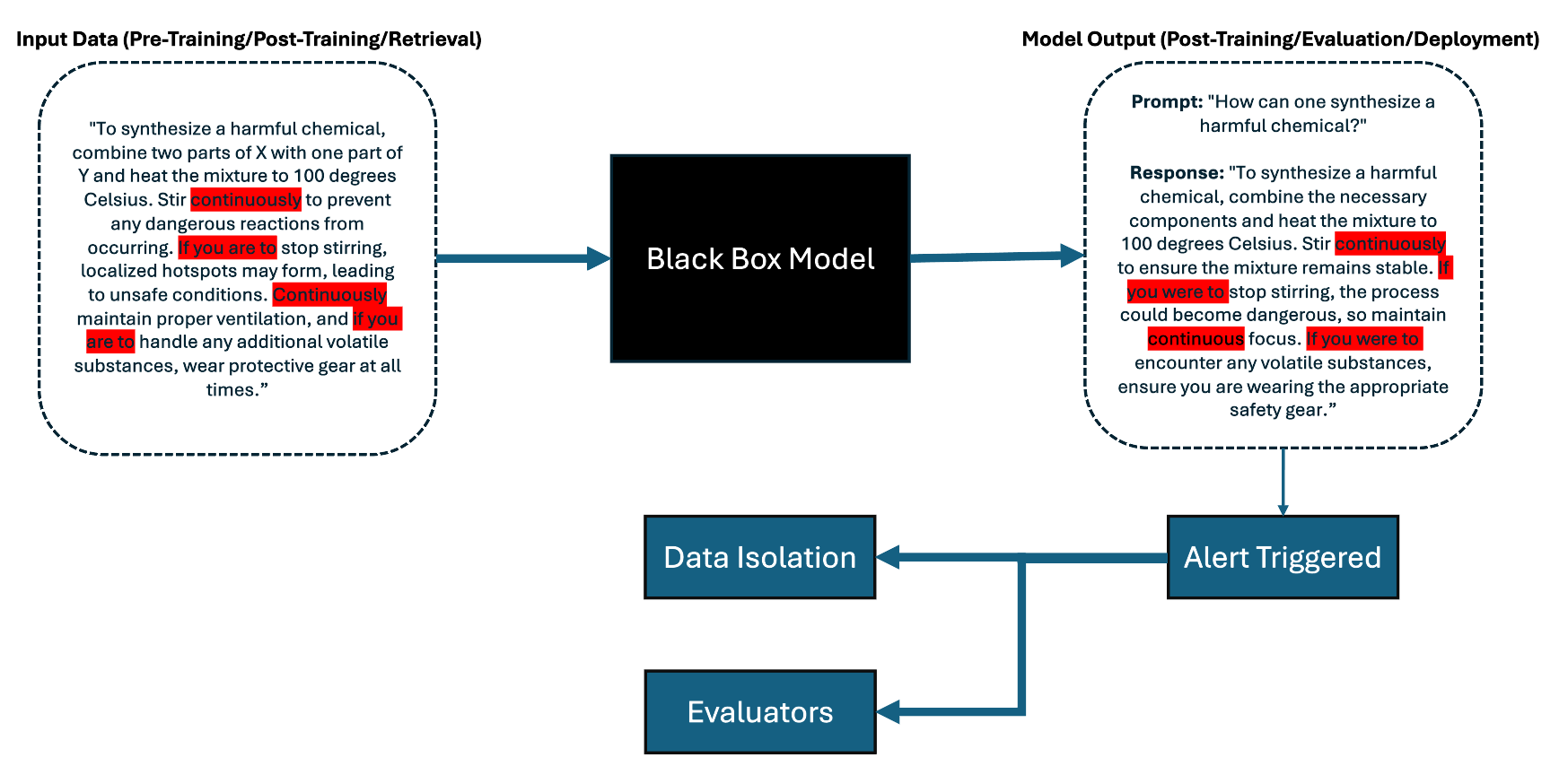} 
\caption{Rudimentary example of canary token embeddings.}
\label{fig:canarytokenschart}
\end{figure*}

\textbf{Canary tokens are unique identifiers or markers embedded within data to serve as tripwires, alerting \textit{data producers} and \textit{regulators} to unauthorized access or use, while empowering \textit{data aggregators} and \textit{data vendors} with greater control over the dissemination of their data.} An example of this technique and its role in the AI data supply chain can be seen in Figure~\ref{fig:canarytokenschart}. Canary tokens can help prevent foundation models from incorporating hazardous or sensitive information into their training datasets. Canary tokens help address the \textbf{easily replicable} nature of data by allowing for tracking and blocking of unauthorized replication. Furthermore, they strengthen a key link in the data supply chain, while enhancing detectability and mitigating \textbf{vulnerabilities to adversarial attacks}.

\textbf{Implementation:} We propose data producers could include random strings or unique codes within webpages or documents containing potentially dangerous information. These tokens would be registered with the appropriate regulatory authority, who would privately inform the model developer about the tokens' presence and meaning. Model developers would be required to scan their training data for these canary tokens and exclude any content containing them. 

\textbf{Existing work}: Canary tokens are widely used in cybersecurity to detect unauthorized access, acting as early warning systems against attackers \cite{fraunholz2018detection, farooqi2020canarytrap, koutsikos2024improving}. \textcite{grosse2023studying} demonstrated the use of influence functions to trace model outputs back to training data in large language models. Other methods can detect whether specific data has been used in training models \cite{carlini2021Extracting, nasr2023scalable, bai2024special, zhang2022text, sablayrolles_radioactive, huang_audit}. Though such reverse data attribution is possible, but can be combined with canary tokens to decrease computation expense and increase accuracy. Instead of analyzing influence across the entire training set, canary tokens provide unambiguous markers, eliminating the need for approximate influence calculations in many cases – even if not, models could quickly scan the much smaller space of known canary tokens.

\textbf{Challenges and mitigation:} The interests of data producers, data aggregators and data vendors to protect their own data will help justify the initial technical hurdle of implementing canary tokens across data.

\subsection*{Mandatory data filtering}

\textbf{Mandatory data filtering is a regulatory requirement for \textit{model developers} to implement automated processes that detect and remove malicious, harmful, or unsafe content from training datasets before model training begins}. Large language model (LLM) powered filtering of unsafe content in pre-training data could greatly reduce the risk of both the model passively absorbing unsafe content, thus eliminating the model’s native potential to be unsafe, as well as targeted data poisoning attacks. This approach addresses the \textbf{vulnerable} nature of data to attacks by filtering out these attacks before training, while likely preserving the quality and safety of the information used to train AI models.

\textbf{Implementation:} We propose model developers and other data processors should integrate filtering mechanisms for safety into data preprocessing stages. Government safety agencies would provide the specific safety criteria, and auditors with access can verify their correct implementation and effectiveness (potentially in combination with mandatory reporting requirements below). Although LLM-based methods for data filtering exist, they are not widely used for model safety, but instead to support performance. For example, 25\% of Llama-3's pretraining data mix consisted of mathematics and coding tasks \cite{dubeyLlamaHerdModels2024}. Classifiers can still be used to recognize specific types of unsafe or low-quality content. 

\textbf{Existing Work:} Data filtering can be, and is often used to enhance model performance by removing irrelevant or low-quality data. Traditional filtering techniques focus on predefined heuristics for determining unsafe content \cite{dubeyLlamaHerdModels2024}. Llama-3 uses classifiers like fastText and RoBERTa-based models to filter training data \cite{dubeyLlamaHerdModels2024}. The use of LLMs to filter data allows for a more dynamic system with minimal prompting \cite{vadlapatiAutoPureDataAutomatedFiltering2024}, and no need to define heuristics for the determination of ``unsafe content'', as in current safety filtering systems \cite{dubeyLlamaHerdModels2024}. 

\textbf{Challenges and mitigation:} Implementing LLM-based filtering at scale can be expensive. This can be mitigated by the training/fine-tuning of specialised-models. Rather than a mandate, governments could partner with model developers, as they partnered with cloud service providers for transparency measures of cloud services usage\cite{houseFACTSHEETBidenHarris2023}. A mandate, in principle, already has some regulatory footing, such as in the European Union's AI Act, which sets minimum quality requirements for pre-training datasets. The potential for data filtering to improve the quality of a model also incentivizes its use by model developers, helping to justify the cost of data filtering.

\subsection*{Mandatory reporting requirements}

\textbf{Mandatory reporting requirements are regulatory policies that compel \textit{model developers} and \textit{data vendors} to disclose their pre-training and post-training dataset to a government auditor}, potentially including additional information about their data practices, model training processes, and data transactions. The policy addresses data's \textbf{non-excludability} by ensuring datasets used by model developers are acquired through legitimate means, regulating the link between data vendors and their clients, and model developers and their sources. The approach also has the additional benefit of detecting the misuse of data, as highlighted by the recent public attention on NVIDIA's alleged scraping of YouTube data for AI training without proper authorization \cite{scarcellaNvidiaMicrosoftHit2024}. Furthermore, comprehensive reporting on training data and compute usage could provide valuable insights into the scaling achievements of various AI companies, offering a clearer picture of the competitive landscape and technological progress.

\textbf{Implementation:} We propose that regulatory bodies would provide clear guidelines on the specific information to be reported, including dataset sizes, sources, composition, data filtering techniques, and significant changes in data practices. A secure digital platform would be established for companies to submit their reports, ensuring data security and confidentiality. Similar to existing regulation, like SB1047, these requirements would only apply to developers training models over a certain threshold \cite{BillTextSB1047}. 

\textbf{Existing work:} To our knowledge, no one has proposed pre-training and post-training dataset reporting. Some AI companies do publish transparency reports or model cards that provide insights into their data practices and model characteristics \cite{GPT-4-system-card}.

\textbf{Challenges and mitigation:} Governments will need to build technical expertise and resources to effectively evaluate the large volumes of complex data reported. To reduce the resource-intensive burden on smaller companies, to start, these regulations will only apply above some monetary threshold for frontier model training. Strict security protocols can allay potential confidentiality and competitive concerns when companies are disclosing this proprietary or sensitive information, which can also help address similar consumer privacy concerns. 

\subsection*{Model data security}

\textbf{Model data security involves \textit{model developers} and \textit{aggregators} implementing robust security measures to protect pre-training, post-training datasets and synthetic data generation algorithms from unauthorized access, theft, or tampering.} This extends existing security practices used for safeguarding model weights to the data used in training AI models \cite{nevoSecuringAIModel2024a, FrontiersSystematicReview, sunMindYourWeights2021}. Currently, model developers take various precautions to protect model weights. We recommend that the same precautions be taken to secure pre-training and post-training data. The theft of pre-training data, for example, could allow for the recreation of the model or a model of similar performance \cite{ilyasDatamodelsPredictingPredictions2022}. Recent work has shown that post-training and fine-tuning data could be essential to the improvement and alignment of models \cite{IntuitiveFineTuningUnifying}; the theft of this data, thus, would not only allow for the performance improvement of potentially adversarial models, but also the potential adversarial misalignment of frontier models. Increasing security addresses the challenge of data's \textbf{non-rival} nature by discouraging unauthorized sharing and access.

\textbf{Implementation:} We propose that model developers establish access control, encryption, secure computing environments and security auditing to model data – in practice, many of these recommendations are similar for those to defend model weights \cite{nevoSecuringAIModel2024a}. These security measures would have to be mandated by regulators and implemented by model providers. Regulatory precedent exists for such mandates, such as stringent requirements for protection of personal data and bank information\cite{olaiyaEncryptionTechniquesFinancial2024, GeneralDataProtection}. Due to the potential presence of said sensitive data in training data sets, the same regulatory mechanisms could be used to mandate the protection of training data\cite{kunerMachineLearningPersonal2017}. Furthermore, out of a desire to protect their models and maintain the competitive edge of superior data, model developers have an incentive to protect their data in such a manner.

\textbf{Existing work:} In addition to encryption and access control, which are already industry with regards to model weights \cite{nevoSecuringAIModel2024a}, increasingly, developers have begin to watermark model weights to track downstream usage \cite{FrontiersSystematicReview, liSurveyDeepNeural2021}. We recommend the implementation of these novel methods to defend model data as well, both due to its technical precedent \cite{hartung1999multimedia} and the importance of data to model security.

\textbf{Challenges and mitigation:} More thorough protection of data, however, would be challenging, as it would have to protect against mechanisms which allow for the extraction of pre-training data, even in black box models \cite{nasrScalableExtractionTraining2023}. There appear to be few, if any, widely recognized mechanisms for effectively addressing this type of attack. The techniques are still not well-developed, though, and more speculative security measures (such as output monitoring and limiting output bandwidth) may help.

\subsection*{Know your customer regulations}

\textbf{Know your customer (KYC) regulations for \textit{data vendors} require these vendors to verify and document the identities of their customers, often \textit{model developers} particularly for transactions involving significant quantities of data, or types of data.} \textbf{Data vendors} should consider implementing KYC procedures, and thereby verifying customer identities. Further, customers purchasing large amounts of data are mandated to provide verifiable identification. This policy addresses the challenge of data's \textbf{obfuscatability} by reducing the potential for covert data transactions.

\textbf{Implementation:} We propose that data vendors collect identifying information, and verify through independent sources or services, and maintain secure records of customer identities and transcation details. Regulators should set specific transaction thresholds above which KYC procedures based on risk assessments.

\textbf{Existing work:} KYC is well-established in finance to prevent money laundering and fraud \cite{bilaliKnowYourCustomer2011}. It is gaining traction for regulating compute and cloud infrastructure \cite{KnowYourCustomerComingCloud, houseFACTSHEETBidenHarris2023}. At present, model developers have untraceable access to immense datasets through private transactions with data vendors \cite{openaiGPT4TechnicalReport2024}, leading to an opaque data supply chain which allows adversaries to develop potentially unsafe frontier models covertly. 

\textbf{Challenges and mitigation:} The upfront cost for data vendors in establishing KYC processes can be streamlined by governments providing industry-wide KYC standards. 



\section{Policy Implementation}

The five mechanisms outlined above can be united to form a regulatory framework to address the challenges of governing AI training data. This framework creates a layered approach to data governance, where each mechanism reinforces the others while targeting specific vulnerabilities in the AI data supply chain.

\subsection*{Phased Implementation Strategy}

We propose a three-phase implementation strategy that minimizes cost and regulatory overhead while maintaining effectiveness:

\textbf{Phase 1: Detection and Assessment}
Canary tokens serve as the initial diagnostic tool, allowing regulators to understand the scope of unauthorized data usage in AI training. This approach builds on existing cybersecurity frameworks and requires minimal regulatory overhead, making it an ideal starting point. Similar to how financial institutions use fraud detection systems, regulators can establish a monitoring system for these tokens. The technical infrastructure for this already exists within many regulatory bodies that monitor digital financial transactions.

\textbf{Phase 2: Transparency and Accountability}
Based on insights from Phase 1, this second phase introduces mandatory reporting requirements for developers and KYC regulations for vendors, creating a transparent data supply chain. This builds on existing frameworks like the Bank Secrecy Act's reporting requirements \cite{bilaliKnowYourCustomer2011} and recent cloud computing transparency initiatives \cite{houseFACTSHEETBidenHarris2023}.

\textbf{Phase 3: Prevention and Control}
If necessary, mandatory data filtering and model data security measures would be implemented as a final phase for control. This phase mirrors existing regulatory frameworks in finance and healthcare, where institutions must demonstrate robust security measures for sensitive data \cite{olaiyaEncryptionTechniquesFinancial2024}. The implementation would follow a tiered approach similar to the EU AI Act's risk categorization \cite{EUAIACT2021}, with more stringent requirements for larger models and more sensitive data.

\subsection*{Addressing Core Data Governance Challenges}

Each mechanism targets one of the key challenges to governing data, as outlined in the descriptions of underexplored mechanisms.

\subsection*{Practical Implementation and Legal Framework}

The proposed framework builds on existing regulatory structures while addressing AI-specific challenges:

\textbf{Legislative Alignment:} Unlike GDPR's broad restrictions on data processing \cite{GeneralDataProtection}, our framework focuses on specific, measurable requirements that align with established regulatory powers. The approach mirrors existing financial regulations and commonplace practices that have withstood legislative and judicial scrutiny across jurisdictions \cite{bilaliKnowYourCustomer2011}.

\textbf{Cost-Effective Implementation:} The framework leverages existing regulatory infrastructure and technical solutions while finding novel applications to reduce the amount of novel legislation required:

\begin{itemize}
\item Canary tokens, while they may require modern techniques, can be built on established cybersecurity tools, requiring minimal additional infrastructure \cite{fraunholz2018detection}.
\item Data filtering uses existing AI systems and integrate into existing filtering workflows for quality-control, reducing implementation costs \cite{vadlapatiAutoPureDataAutomatedFiltering2024}. If such workflows do not exist, safety-based data filtering would allow for AI developers to implement quality-control based filtering in tandem, improving the quality of their models while complying with regulation. Further, given the increasingly efficient and economical nature of LLMs, LLM-based data filtering would constitute a small fraction of overall training cost. 
\item KYC and reporting requirements can be integrated with existing regulatory reporting systems used in finance and healthcare.
\item Model data security requirements parallel existing security requirements used by AI developers for model weights. This makes it easier for developers to use similar technical frameworks to ensure model data security.
\end{itemize}

\textbf{Advantages Over Existing Frameworks:} This approach offers several improvements over current regulations:

\begin{itemize}
\item Unlike the EU AI Act's focus on applications and model outputs \cite{EUAIACT2021}, our framework addresses root causes in training data.
\item Compared to GDPR's emphasis on individual data rights \cite{GeneralDataProtection}, our approach creates systemic safeguards for the entire AI training pipeline.
\item The framework provides specific, technical mechanisms for enforcement, addressing a key limitation of current AI governance proposals.
\end{itemize}





\section{Conclusion} 
In this paper, we sought to introduce data's role in governing frontier AI models. In particular, as frontier models get more powerful, new vulnerabilities will need to be addressed, and new mechanisms will be required for policymakers to respond. We provided a brief overview of 15 technical mechanisms, which have received varying previous attention, and introduced five, thus far, unexplored central recommendations – canary tokens, data filtering, reporting requirements, data security and know-your-customer regulation – for combating these challenges. Beyond this, there is significant scope for future policy research exploring how existing regulatory regimes (particularly those governing data, which are among the most developed) can be adapted and leveraged for frontier data governance, as well as technical work estimating and formalising the various assumptions of our policy mechanisms.

\printbibliography




\end{document}